\documentclass{article} 
\usepackage{iclr2015,times}
\usepackage{hyperref}
\usepackage{url}
\usepackage{amsmath}
\usepackage{amsfonts}
\usepackage{amssymb}
\usepackage{enumerate}
\usepackage{color}
\usepackage{caption}
\usepackage{subcaption}
\usepackage{multirow}

\usepackage{tikz}
\usetikzlibrary{shapes,arrows}

\tikzstyle{block} = [rectangle, draw, fill=gray!10, 
    text width=5em, text centered, minimum height=4em] 
\tikzstyle{line} = [draw, -latex']
\tikzstyle{cloud} = [draw, rectangle,fill=gray!20, node distance=2em,
    minimum height=4em]

\title{Convolutional Neural Networks for joint \\ object detection and pose estimation: \\ A comparative study}

\author{
Francisco Massa, Mathieu Aubry, Renaud Marlet\\
Universit\'e Paris-Est, LIGM (UMR CNRS 8049), ENPC\\
 F-77455 Marne-la-Vall\'ee, France\\
\texttt{francisco-vitor.suzano-massa@polytechnique.edu} \\
\texttt{mathieu.aubry@imagine.enpc.fr} \\
\texttt{renaud.marlet@enpc.fr}
}

\newcommand{\pos}{\mathit{pos}}
\renewcommand{\neg}{\mathit{neg}}
\newcommand{\pose}{\mathit{\,pose}}
\newcommand{\class}{\mathit{\,class}}

\definecolor{dkmag}{rgb}{0.5,0,0.5}
\definecolor{dkblue}{rgb}{0.0,0,1.0}

\newcommand{\change}[1]{{{#1}}}
\newcommand{\checck}[1]{{{#1}}}

\iclrconference 

\begin{document}

\maketitle

\begin{abstract}
In this paper we 
study the application of
convolutional neural networks for jointly detecting objects depicted
in still images and estimating their 3D pose. 
We identify different feature representations of oriented objects,
and energies that lead a network to learn this representations. The choice of the
representation is crucial since the pose of an object has
a natural, continuous structure while its category is a discrete variable. 
We evaluate the different approaches on the joint object detection
and pose estimation task of the Pascal3D+ benchmark using Average Viewpoint Precision. 
We show that 
a classification approach on discretized viewpoints achieves state-of-the-art performance for
joint object detection and pose
estimation,
and significantly outperforms existing baselines on this benchmark. \change{We also show that performing the two tasks jointly can improve significantly the detection performances.}
\end{abstract}

%

\section{Introduction}
Convolutional Neural Networks (CNNs) have long shown impressive and practical results on specialized vision tasks, such as optical character recognition \citep{lecun1998gradient,simard2003best}. Thanks to the availability of large, annotated image datasets \citep{deng2009imagenet} and the increased power of GPU computing, CNNs have recently outperformed other techniques in less constrained computer vision tasks. Following the seminal work of \citet{krizhevsky2012imagenet} on ImageNet classification many methods have been developed to apply CNNs to other tasks. 
In this work, we explore the potential of CNNs in a specialized task on real world images, namely joint detection and pose estimation in images from the difficult Pascal VOC dataset \citep{pascal-voc-2012} extended with 3D annotations \citep{xiang2014beyond}.

Detecting and estimating the pose of a known object instance is one of the oldest problem of computer vision \citep{roberts1963machine}. In early works, object contours were detected explicitly in images and matched to the contours of a 3D model of the object instance \citep{mundy2006object}. As the field moved towards object category recognition, methods focused on photographs of the objects and performed detection without explicitly taking into account the 3D structure of the object. These approaches are typically based on manually designed low-level features, such as HOG \citep{dalal2005histograms} and SIFT \citep{lowe2004distinctive}. These features are then aggregated and used as the input of a classifier such as a SVM. Recently, the problem of pose estimation for object categories has drawn attention, and many techniques have been developed (see section \ref{sec:related_work} for an overview).

Joint object detection and orientation estimation in real images is difficult because it requires features with two conflicting properties. 
First, estimating orientation requires a high level of discriminative power, since for example the rotation of an airplane by 10 degrees will not change its appearance much. Second, the features need to cope with 
strong appearance changes since objects may be occluded or their appearance may change, e.g.\ because of the intra-class variability of the categories or illumination effects. 
Moreover, the amount of training data for this task is very limited because of both the difficulty of annotating object orientation and the fact that some viewpoints are less common (for example a TV monitor is rarely seen from behind).

To deal with the lack of training data, following \citet{Oquab13} and \citet{kaiming14ECCV}, we use a CNN pre-trained on ImageNet \citep{deng2009imagenet} and only learn or fine-tune the last layers. Based on this idea, we choose to use the Spatial Pyramid Pooling (SPP) framework \citep{kaiming14ECCV}. It consists of computing densely over the image the convolutional layers of a pre-trained CNN in a manner similar to \citet{sermanet2013overfeat}, and re-scaling the output to compute the fully connected layers which are fine-tuned for the task. This framework allows efficient training and testing by taking advantage of shared computation in the first layers, and provides state-of-the-art performance in both classification and detection. 
We adapt and fine-tune the last layers of the SPP architecture to predict different features which identify both the presence of an object and its orientation. In all our experiments, the same network performs both detection and pose estimation, the only difference being the feature used and the associated \change{error function}.

Our goal can be related to an emerging trend in computer vision, which attempts to predict more detailed information about objects in addition to their category, such as 3D information. To achieve this goal, the question of the representation of an object category, which is not seen as a single entity anymore, is central. For the special case of orientation, we identify and test several natural representations.

\paragraph{Contributions.} In this paper, we compare 
 several feature representations, learned with a single CNN, for joint detection and pose estimation. We show that handling the continuous nature of object poses is \change{possible}, but that a method which discretizes object poses and classify each image patch as both an object and a pose performs well. Moreover it significantly outperforms existing methods, providing a new baseline for the task. We also demonstrate the benefit of continuous representations

\section{Related work}
\label{sec:related_work}
\paragraph{Pose estimation.} Computer vision work focusing on pose estimation can be roughly separated in two classes: those which aim exclusively at instance alignment and those which target alignment for an entire category. Work on instance alignment includes both very early results such as \citep{huttenlocher1990recognizing,lowe1987three} and more recent ones aiming at difficult or specialized tasks \citep{lim2013parsing,aubry2014seeing}. To perform category-level pose estimation, it is necessary to handle the large intra-class variation of the categories. That may be difficult to do in 3D, and for this reason most work either focuses on classes with a simple 3D geometry, such as cars, or consider simplified models of objects, for example approximated with planes \citep{xiang2012estimating} or cuboids \citep{xiao2012localizing,fidler20123d}. These hypotheses clearly limit the applicability of the methods for general objects and real images.

Most of these approaches were demonstrated only on a restricted dataset. Recently, \citet{xiang2014beyond} introduced an extension of the classic and challenging Pascal VOC dataset \citep{pascal-voc-2012} by aligning a set of CAD models for 12 object classes. They also introduced the Average Viewpoint Precision as a standard metric for the evaluation of joint detection and viewpoint estimation. They provided baseline evaluations using a DPM detector for each orientation class as well as using the method of \citet{pepik2012teaching} which uses an adaptation of DPM with 3D constraints and usually performs better. We show that we can clearly outperform this baseline using CNNs.

\paragraph{CNNs.} Convolutional neural networks \citep{lecun1989backpropagation} are architectures designed to learn robust features over images. They are formed by a succession of layers such as convolutions or spatial poolings. They have been successfully applied to many specialized tasks, such as digit recognition  \citep{lecun1998gradient,simard2003best} or, more related to our work,  joint detection and pose estimation for faces \citep{osadchy07}.  However they have attracted much more attention in computer vision after the impressive advances made on the classification task for the 1000 categories of ILSVRC 2012 by \citet{krizhevsky2012imagenet}. Since then, they improved dramatically results on many vision problems, being either adapted to the specific task or used as generic features \citep{sermanet2013overfeat}.

Object detection is of special interest to us and can be easily achieved by classifying a set of bounding boxes provided by a selective search algorithm \citep{van2011segmentation}. This strategy was first used by \citet{girshick2013rich} and led to a dramatic improvement of Pascal VOC detection results. The drawback of this algorithm is that the CNN must be applied independently to each candidate bounding box. For this reason, \citet{kaiming14ECCV} proposed to apply selective search after the convolutional layers, leading to a more efficient algorithm and achieving slightly better performance.

Many methods have recently been proposed to use neural networks with less training data than is available in ImageNet. In particular \citet{Oquab13} performed classification on Pascal VOC dataset by adding and learning two fully connected layers on top of a network already trained on ImageNet, and \citet{kaiming14ECCV}  obtained state-of-the-art results in both detection and classification by learning new fully connected layers on top of convolutional layers trained on ImageNet. It remains however unclear how much invariance is encoded in the first layers of the network, and if such a strategy could be applied to a task such as pose estimation, which requires to be more discriminative.





\section{Predicting orientation with Convolutional Neural Networks}
\label{sec:unif}

\begin{figure*}[t]
\centering
\newcommand{\imgheight}{0.35\linewidth}
\begin{subfigure}[b]{0.22\linewidth}
\includegraphics[width=0.95\linewidth]{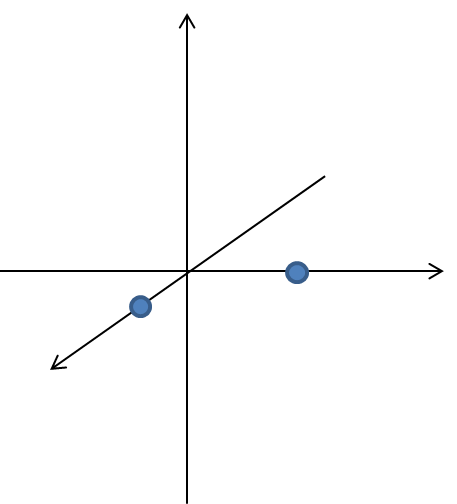}
\caption{Section \ref{sec:discrete}}
\label{fig:41}
\end{subfigure}
\hfill\hfill
\begin{subfigure}[b]{0.22\linewidth}
\includegraphics[width=0.95\linewidth]{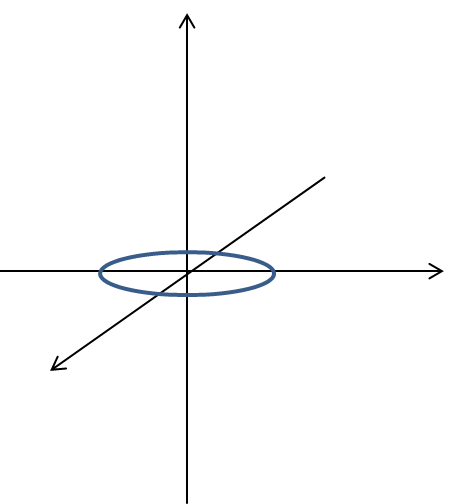}
\caption{Section~\ref{sec:manifold}}
\label{fig:42}
\end{subfigure}
\hfill
\begin{subfigure}[b]{0.45\linewidth}
\hfill
\raisebox{0.5mm}{\includegraphics[width=0.45\linewidth]{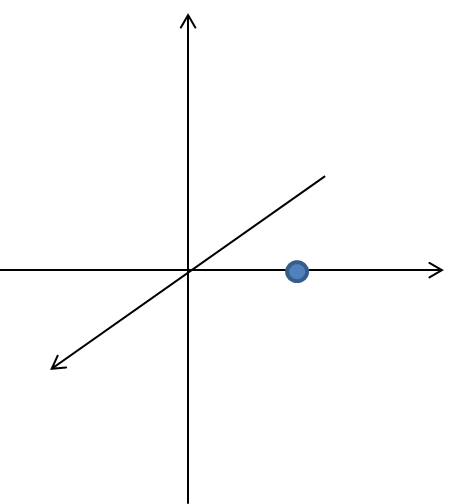}}
\raisebox{8mm}{\includegraphics[width=0.45\linewidth]{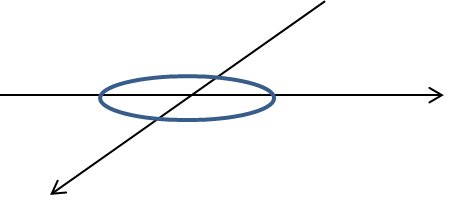}}
\caption{Section~\ref{sec:regression}(a)}
\label{fig:43a}
\end{subfigure}
\\
\begin{subfigure}[b]{0.45\linewidth}
\vspace{2mm}
\hfill
\includegraphics[width=0.45\linewidth]{point}
\includegraphics[width=0.45\linewidth]{circle3d}
\hfill
\caption{Section~\ref{sec:regression}(b1)}
\label{fig:43b1}
\end{subfigure}
\hfill
\begin{subfigure}[b]{0.45\linewidth}
\includegraphics[width=0.45\linewidth]{point}
\includegraphics[width=0.45\linewidth]{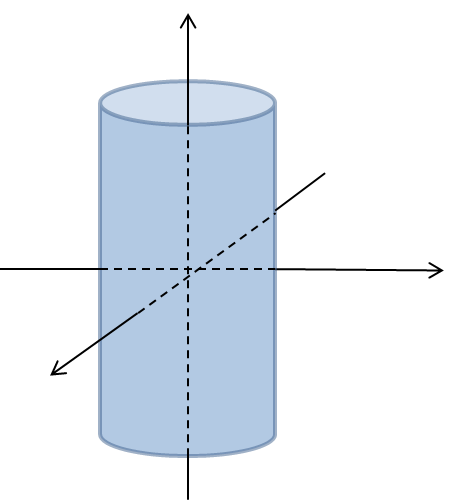}
\hfill
\caption{Section~\ref{sec:regression}(b2)}
\label{fig:43b2}
\end{subfigure}
\vspace{-2mm}
\caption{Feature representation for \change{the} joint detection and pose estimation \change{methods} explored in this paper. \change{For each approach,} the features associated to the different orientations of a given object category \change{correspond to a subspace of some general feature space. In the above images, this target subspace for one object class and all its orientations is shown in blue in a 3D subspace of the actual feature space. For the discrete method network (section~\ref{sec:discrete}), the subspace is a set of discrete points corresponding to a combination of both a category and a discrete pose (only two poses shown here). For the single-class regression (section~\ref{sec:manifold}), this subspace is a 2D circle in a 3D space. For the joint classification and continuous regression (section~\ref{sec:regression})}, we plotted independently the classification features (left\change{: discretized points}) and \change{the} pose \change{features} (right\change{: 2D circle in 2D space, 2D circle in a higher dimension space,  or cylinder}).}
\vspace{-3mm}
\label{fig:feat}
\end{figure*}

Except when stated otherwise, we consider the problem of detection and pose estimation for $N$ object classes. Given that the pitch and roll angles vary only slightly for Pascal VOC images, we restrict ourselves to the estimation of the azimuth angle in this paper. However, our model could be extended to account for other angles.

The key elements to define in order to learn a CNN that predicts both object class and pose are:
\begin{enumerate}
\item a feature space where each object and pose is associated to a given point or set of points,
\item a cost function defining the loss for each prediction to be minimized during training,
\item a network architecture.
\end{enumerate}

The choice of features sometimes seems obvious, for example having a $N$-dimensional vector as output to classify an image into $N$ discrete classes. However, it is not the case for our problem, because our output has a continuous structure. For example, it seems more natural to represent the features associated to chairs of different orientations as a circle rather than a set of separated points.  For this reason, we tested several alternative representations, which are summarized in figure \ref{fig:feat}.

In section \ref{sec:discrete} we adapted directly the successful algorithms developed for detection by discretizing the different object poses into $P$ orientations for $N$ different classes, predicting a $N\times P+1$ vector with the probabilities for each class and orientation\change{, plus the background class}.

On the contrary, in section \ref{sec:manifold} we focus on the continuous aspect of angle prediction. For a single class, we associate to each orientation of the object a point on a circle, and force background patches to have features far from this circle. This way, detection and pose estimation are completely joined: small distances from the features to the circle are associated to the class, and the angle predicts orientation.

Section \ref{sec:regression} can be seen as an intermediate between the two previous approaches. For each patch, we predict jointly the probability of the class as a $N\times1$ vector and the angle. The manifold corresponding to the angle feature can be a single circle for all classes, a circle per class, or an hyper-cylinder per class: we evaluate each option.

The \change{error function} we use for each case is different and adapted to the feature we want to learn, our choices for each approach are detailed in section \ref{sec:select}.

Finally, the choice of the CNN architecture has of course an important influence on the results. To provide comparable results, we choose to base all our models on the Spatial Pyramid Pooling framework developed by \citet{kaiming14ECCV}, which is efficient for training and testing. \change{We used the "Zeiler5" architecture which provides good results for detection.} Dense features are computed over the image at multiple scales using five convolutional layers and the two max-pooling layers similar to those of \citet{krizhevsky2012imagenet}. This first part of the network is trained on ImageNet and we used the network provided by \citet{kaiming14ECCV}. The features corresponding to windows picked up by selective search over the image  \citep{van2011segmentation} are then rescaled and used as input for the \change{three} fully-connected layers.  As in the work of \citet{kaiming14ECCV}, we fine tune only the fully-connected layers for our different tasks.


\section{Selected frameworks for pose estimation}
\label{sec:select}
In this section, we describe in more details the different approaches that are evaluated in this paper and the corresponding choices of \change{error functions}. They were selected to cover the most different approaches to the problem, corresponding to the different cases presented in figure \ref{fig:feat}.



In all the section, we write  $y(\mathbf{x},\mathbf{w})$ the output of a network $y$ of parameters $\mathbf{w}$ for the input $\mathbf{x}$. We suppose we are given $M$ training images $(\mathbf{x}^k)_{1\leq k\leq M}$, with ground truth labels $\mathbf{t}^k$ and orientation $\theta^k$ for each~$k$. As explained in section \ref{sec:unif}, labels $\mathbf{t}^k$ have different dimensions and interpretations depending on the choice of architecture.

\subsection{Discretized pose classification}
\label{sec:discrete}
The simplest way to solve the problem of pose estimation based on a classification algorithm is to discretize the orientations into $P$ bins and to consider each bin as a separate class. This leads to a classification problem with $N\times P+1$ classes. We write $(t_0,t_{1,1}, t_{1,2},...,t_{1,P},t_{2,1},...,t_{N,P})$ the coordinates of a vector $\mathbf{t}$ in $\mathbb{R}^{N\times P+1}$. We define $\mathbf{t}^k= (t^k_0,t^k_{1,1},t^k_{1,2},...,t^k_{N,P})$ by setting $t^k_{i,j}=1$ if $\mathbf{x}^k$ is of object category $i$ and orientation $j$ and 0 otherwise, setting $t^k_0=1$ if the object is not from any category, and 0 otherwise.  We use exactly the same SPP architecture as \citet{kaiming14ECCV} for detection. The first part of the network trained on ImageNet is fixed. Windows corresponding to patches extracted by selective search on the image are extracted from the feature maps and rescaled to a fixed size input to three fully connected layers, followed by a softmax. The \change{error} minimized by the learning algorithm is the negative log-likelihood, which can be written:
\begin{equation}
E(\mathbf{w})=-\sum_{k=1}^M \left( t^k_0\log \left(y(\mathbf{x}^k,\mathbf{w})\right)_0 + \sum_{i=1}^N \sum_{j=1}^P t^k_{i,j}\log \left( y(\mathbf{x}^k,\mathbf{w})\right)_{i,j}\right)
\label{eq:discrete}
\end{equation} 
Because of the softmax, $y(\mathbf{x},\mathbf{w})$ has all its coordinates between 0 and 1 and sums to 1. Thus the \change{error} is minimal for $y(\mathbf{x}^k,\mathbf{w})=t^k$.
For a test input $\mathbf{x}$, $\left(y(\mathbf{x},\mathbf{w})\right)_{i,j}$ can be interpreted as the probability for the input to be of class $i$ and orientation $j$.


\subsection{Continuous regression}
\label{sec:manifold}
The discrete framework does not consider the natural continuous structure of the object space. Indeed, the appearance varies continuously with the viewpoint while the output of the discretized network is supposed to ``jump'' from one point to another when the viewpoint varies, not taking advantage of the continuity of the appearance. For this reason, we present in this section a different view of the problem, similar to the one developed by \citet{osadchy07} for face detection and orientation prediction. We consider a single class and 
we associate to each positive example a point on the unit circle, with its angle corresponding to the orientation. We do not enforce any specific prediction for the negative examples, but we force them to be far away from the circle describing the positives.
A problem that arises if one simply considers a circle in the 2D space is that an \change{error function} defined to push symmetrically the negatives away from the circle will have a local minimum inside the circle for the negatives. Thus, some negative samples may be trapped into the circle instead of being pushed far away. To avoid this effect, we add a third dimension to our features $\mathbf{t}$ and let the 2D unit circle live in a 3D space. More formally, the circle defining the positives is defined as $\chi=\{(\cos(\theta),\sin(\theta),0), \theta\in \mathbb{R\}}$. Since the problem is not a classification problem, we use the network of \citet{kaiming14ECCV}, without the final softmax.

We separate the indices of the training examples into the sets $\mathcal{S}_\pos$ corresponding to positive examples and the set $\mathcal{S}_\neg$ corresponding to negative examples. For $k\in\mathcal{S}_\pos$, the desired output of the network is $\mathbf{t}^k=(\cos(\theta^k),\sin(\theta^k),0)$. 
An \change{error function} that attracts positive examples to their target position on the network can be written as:  
\begin{equation}
E(\mathbf{w})= \dfrac{1}{|\mathcal{S}_\pos|}\sum_{k\in \mathcal{S}_\pos} L^\pos(\mathbf{x}^k,\mathbf{t}^k,\mathbf{w}) + K  \dfrac{1}{|\mathcal{S}_\neg|} \sum_{k \in \mathcal{S}_\neg}L^\neg(\mathbf{x}^k,\mathbf{w})
\end{equation} 
where $K$ weighs the respective contributions of the positive and negative losses $L^\pos,L^\neg$ in the global \change{error}. The most natural properties to require from the losses $L^\pos$ and $L^\neg$ are that:
\begin{itemize}
 \item $L^\pos(\mathbf{x},\mathbf{t},\mathbf{w})$ should have a local minimum only if $y(\mathbf{x},\mathbf{w})=\mathbf{t}$.
 \item $L^\neg(\mathbf{x},\mathbf{w})$ should be a decreasing function of $\| y(\mathbf{x},\mathbf{w})-\pi(y(\mathbf{x},\mathbf{w})) \|$\change{, where $\pi$ is the projection on $\chi$.}
 \end{itemize}
Following \citet{osadchy07}, we choose the following losses, with the above properties:
\begin{equation}
 L^\pos(\mathbf{x},\mathbf{t},\mathbf{w})=\| y(\mathbf{x},\mathbf{w})- \mathbf{t}\|_2^2
 \label{eq:pose}
\end{equation} 
\begin{equation}
 L^\neg(\mathbf{x},\mathbf{t},\mathbf{w})=\exp\left(\frac{-\| y(\mathbf{x},\mathbf{w})- \pi(y(\mathbf{x},\mathbf{w}))\|_2}{\delta}\right)
\end{equation} 
The parameter $\delta$ was not introduced by \citet{osadchy07} but we found it important and complementary to $K$. Indeed, huge values of $K$ would be needed to compensate the exponential decrease of the negative weights, and push the negative examples at a distance clearly larger than the radius of the positive circle.
With this network, the probability of a test sample $\mathbf{x}$ to correspond to a positive example is a decreasing function of $\| y(\mathbf{x},\mathbf{w})-\pi(y(\mathbf{x},\mathbf{w})) \|$ and its most probable orientation if it is a positive sample is the angle defined by its projection on $\chi$. \change{From a computed feature $y(\mathbf{x},\mathbf{w})$, we obtain the angle corresponding to the pose by $\tilde{\theta^k} = \arctan(y(\mathbf{x},\mathbf{w})_1,y(\mathbf{x},\mathbf{w})_2)$, where $y(\mathbf{x},\mathbf{w})_i$ is the $i$-th coordinate of feature $y(\mathbf{x},\mathbf{w})$.}

\subsection{Joint classification and continuous pose estimation}
\label{sec:regression}
The network described in the previous paragraph treats the classification problem implicitly and is not straightforward to extend to several mutually exclusive classes. One would need for example to define parameters such as the ratio of the distance between the categories and the distance between different orientations of objects of the same category. For this reason, we considered an alternative inspired by the work of \citet{penedones11}, which consists of dividing the last layer of the network in two, one part devoted to the classification that we call $y^\class$ and the other to orientation prediction, $y^\pose$. The classification layer is followed by a softmax while the pose prediction layer is not. For the rest of the network, we keep the same architecture as in the previous sections. This approach can be seen as an intermediate between the other two.

The target values for this network output can be decomposed as $\mathbf{t}=(\mathbf{t}^\class,\mathbf{t}^\pose)$, each part corresponding to the associated last layer. As in a standard classification network, $\mathbf{t}^\class$ is a $N+1$ dimensional vector, each dimension corresponding to the probability that the input is an instance  of the corresponding class. For $\mathbf{t}^\pose$, two natural choices are possible:
\begin{enumerate}[(a)]
\item $\mathbf{t}^\pose$ is a point in 2D whose angle is associated to the viewpoint of an object of any class. 
\item $\mathbf{t}^\pose$ is a point in a $2N$ dimensional space, the angle of the pair $(t^\pose_{2i-1},t^\pose_{2i})$ corresponding to the angle prediction for an object of the object class $i$.
\end{enumerate}
The training labels associated to the first case are $\mathbf{t}^{k,\pose}=(\cos(\theta^k),\sin(\theta^k))$ for the positive examples and the negative examples \change{are not constrained to have any value}. For the second case, \change{$(t^{k,\pose}_{2i-1},t^{k,\pose}_{2i})=(\cos(\theta^k),\sin(\theta^k))$ if $t^{k,\class}_i=1$ and $(0,0)$ if not.}

We define the 
\change{error} of the network by the sum of a classification term and a pose estimation term:
\begin{equation}
E(\mathbf{w})=\lambda E^\class(\mathbf{w})+ E^\pose(\mathbf{w})
\end{equation} 
where $\lambda$ balances the contribution of the two terms. We choose for $E^\class(\mathbf{w})$ the log-likelihood similarly to equation \ref{eq:discrete} using the classification feature. 
%
In case (a), the definition of $E^\pose(\mathbf{w})$ is based on a loss similar to equation \ref{eq:pose}. 
In case (b), 
we consider two possibilities to define $E^\pose(w)$: (b1) penalizes the distance to $\mathbf{t}^\pose$; (b2) penalizes only the distance of the two components corresponding to the relevant class. (b1) is similar to the approach of \citet{penedones11}. \change{An approach similar to (b2) was used in \cite{williams1997instantiating} for control points localization.}

Since performing jointly detection and pose estimation requires optimizing $\lambda$, we provide experiments on pose estimation only using $\lambda=0$\change{, using the detection score of a reference classifier and only learning the pose regression}, and compare several variants of penalization for $L^\pos$. We experimented with the $L^1$, $L^2$ and squared $L^2$ losses.

With this network, the probability of a test sample to belong to each object class is given by the classification part of the output, while the orientation prediction can be made from the pose estimation part in a way similar to section \ref{sec:manifold}.



\section{Experiments}

\subsection{Dataset, evaluation measure and baselines}
We trained and evaluated our methods using the Pascal3D+ dataset introduced by \citet{xiang2014beyond}, which extends the annotation of the challenging Pascal VOC 2012 dataset \citep{pascal-voc-2012}. In particular it provides orientation annotations for 12 object classes. The orientation annotation were obtained by aligning CAD models to the images. \citet{xiang2014beyond} also introduced a standard measure for evaluating joint detection and pose estimation and provided several baselines.

Their accuracy measure, called Average Viewpoint Precision (AVP), is very similar to the standard Average Precision (AP) used for detection and obtained by computing the area under the precision-recall curve, except that it considers a detection to be positive only if the associated orientation prediction is accurate with an error below a given threshold. In the case of discrete viewpoint prediction, the threshold can be associated to a bin size, and thus to the number of different viewpoints that are predicted. For this reason, following \citet{xiang2014beyond}, we report our results with tolerance thresholds corresponding to 4, 8, 16 and 24 views. \change{Since the test set orientation annotations are not available, all the networks were finetuned using only the Pascal training set, augmented with the flipped images, and evaluated on Pascal validation set, as in \citet{xiang2014beyond}. We did not use the data from Imagenet for fine-tunning (except in a single comparison experiment). Moreover, we considered all instances for evaluation, except those marked as difficult; in particular, occluded and truncated objects are evaluated.}

The best baseline provided by \citet{xiang2014beyond} corresponds to the method of \citet{pepik2012teaching}. 
To this baseline, we add two others which consist \change{first} in using the detection method of \citet{kaiming14ECCV} \change{with the same ``Zeiler5'' architecture as used in the rest of the paper. Then, after detection, either} 
(1)~we predict the most probable viewpoint for each class, or (2)~we predict the orientation for each class using a linear regression \change{on the fc7 features computed for each bounding box}. \checck{Note that baseline (2) corresponds to the limit of the method (b2) described in section \ref{sec:regression} when $\lambda$ tends to $+\infty$}. The corresponding results are provided in tables \ref{tab:base} and \ref{tab:baselines}. Detailed results for the CNN baselines can be found in the appendix.
As expected, the AP results are clearly better using CNNs, but one can notice that the AVP is still in favor of the method of \citet{pepik2012teaching}. When comparing baselines (1) and (2), we see that the regression on the fc7 features is not able to recover any valuable orientation information.

\subsection{Training details}
\change{All the networks were trained using Stochastic Gradient Descent, with a momentum of 0.9, a weight decay of 0.0005, and a batch size of 128 patches.}
\change{We balanced the proportion of positive and negative patches in the following manner: for the discretized method (sec.~\ref{sec:discrete}) and for the joint classification and regression models (sec.~\ref{sec:regression}), we use 32 positive patches and 96 negative patches per batch; for the continuous regression (sec.~\ref{sec:manifold}), as there is only one positive class, we consider 8 positive and 120 negative patches per batch.}
\change{We used an adaptive strategy to vary the learning rate and the number of required iterations in our different experiments. For this,}
we \change{randomly} selected a validation set \change{of 6400 patches, keeping the same positive and negative proportions as during training} 
(to be compared to the 127K positive and 13M negative patches generated by the selective search) on which we evaluate every 500 iterations.
\change{We started with a learning rate of $10^{-3}$ (except for the continuous regression case for which we used $5 \times10^{-5}$ to avoid divergence). 
We then divided the learning rate by~2 when the error on our small validation set stopped decreasing for 10 successive evaluations (5K iterations).} 
\change{This strategy allowed a fast training of our networks, although it is not optimal.}
All the models were implemented using Torch7~\citep{torch7_NIPSworkshop2011}.

\subsection{Results}


\paragraph{Discrete method.} \change{For experiments with the discretized method (section~\ref{sec:discrete}), we trained a different model for each discretization variant (4, 8, 16 or 24 poses).} The results 
are presented in table~\ref{tab:discrete}. They clearly outperform the previous state-of-the-art results presented in table~\ref{tab:base}, both in term of AP and AVP. We can see that the detection results are 
\change{comparable}
to the CNN baseline presented in table 
\ref{tab:baselines}
and that they decrease with the number of viewpoints.  Another interesting fact is that the margin by which  our method outperforms the baseline decreases when the number of viewpoints augments. It can be explained by the fact that less training examples per orientation class are available when the number of discrete viewpoints is increased. \change{To test this interpretation, we trained the 24-views model using both Pascal training and Imagenet annotations. The results are presented in the last row of Table~\ref{tab:discrete}. As expected, adding more training data increases considerably the mAP and mAVP.}

\begin{table}[t!]
  \caption{Baseline - DPM-VOC+VP \citep{pepik2012teaching} - [AP\,$|$\,AVP]}
  \vspace{-2mm}
  \label{tab:base}
  \centering
  \setlength{\tabcolsep}{2.3pt}
  {\tiny
  \begin{tabular}{|c|r|r|r|r|r|r|c|r|r|r|r|r|r|r|r|r|r|r|r|r|r|r|r||r|r|}
  \cline{2-26}
   \multicolumn{1}{c|}{}       &\multicolumn{2}{c|}{aeroplane}& \multicolumn{2}{c|}{bicycle} & \multicolumn{2}{c|}{boat} & \multicolumn{1}{c|}{bottle} & \multicolumn{2}{c|}{bus} & \multicolumn{2}{c|}{car} & \multicolumn{2}{c|}{chair} & \multicolumn{2}{c|}{diningtable} & \multicolumn{2}{c|}{motorbike} & \multicolumn{2}{c|}{sofa} & \multicolumn{2}{c|}{train} & \multicolumn{2}{c||}{tvmonitor} & \multicolumn{2}{c|}{Avg.} \\ \hline
4V    	&41.5\relax&37.4	&46.9\relax&43.9&	0.5\relax&0.3&	--&	51.5\relax&48.6&	45.6\relax&36.9&	8.7\relax&6.1&	~~~5.7\relax&2.1&	34.3\relax&31.8&	13.3\relax&11.8&	16.4\relax&11.1&	32.4\relax&32.2&	27.0\relax&23.8\\ \hline
8V	&40.5\relax&28.6&	48.1\relax&40.3&	0.5\relax&0.2&	--&	51.9\relax&38.0&	47.6\relax&36.6&	11.3\relax&9.4&	5.3\relax&2.6&	38.3\relax&32.0&	13.5\relax&11.0&	21.3\relax&9.8&	33.1\relax&28.6&	28.3\relax&21.5\\ \hline
16V	&38.0\relax&15.9&	45.6\relax&22.9&	0.7\relax&0.3&	--&	55.3\relax&49.0&	46.0\relax&29.6&	10.2\relax&6.1&	6.2\relax&2.3&	38.1\relax&16.7&	11.8\relax&7.1&	28.5\relax&20.2&	30.7\relax&19.9&	28.3\relax&17.3\\ \hline
24V	&36.0\relax&\hphantom{0}9.7&	45.9\relax&16.7&	5.3\relax&2.2&	--&	53.9\relax&42.1&	42.1\relax&24.6&	8.0\relax&4.2&	5.4\relax&2.1&	34.8\relax&10.5&	11.0\relax&4.1&	28.2\relax&20.7&	27.3\relax&12.9&	27.1\relax&13.6\\ \hline
  \end{tabular}
  }


  \vspace{3mm}
  \caption{Baselines - CNN. We trained the network for detection only and predicted for the orientation (1) the most probable viewpoint (first line) (2) used a linear regression on fc7 features to predict the viewpoint. }
  \vspace{-2mm}
  \label{tab:baselines}
  \centering
  \setlength{\tabcolsep}{5.7pt}
  {\tiny
  \begin{tabular}{| l | c | c | c | c | c |}
    \hline
     \multicolumn{1}{| c |}{\multirow{2}{*}{\textbf{Baseline}}}   & \multirow{2}{*}{\textbf{mAP}} & \multicolumn{4}{c|}{\textbf{mAVP}} \\
     \cline{3-6} 
         &  & 4V & 8V & 16V & 24V \\
    
    \hline
     \multicolumn{1}{ | l | }{Most probable viewpoint}  & 40.8     & 23.6 & 15.6 & 9.4 & 8.0 \\ \hline
     \multicolumn{1}{ | l | } {Regression on fc7 features}  & 40.8 & 23.9 & 14.9 & 8.7 & 6.9 \\ \hline
  \end{tabular}
  }
  \vspace{3mm}
  \caption{Discretized Method - section \ref{sec:discrete}. The network predicts each orientation as a different class. The last row corresponds to the model trained using Pascal training set + Imagenet annotations - [AP\,$|$\,AVP]}
  \vspace{-2mm}
  \label{tab:discrete}
  \centering
  \setlength{\tabcolsep}{2pt}
  {\tiny
  \begin{tabular}{|c|r|r|r|r|r|r|r|r|r|r|r|r|r|r|r|r|r|r|r|r|r|r|r|r||r|r|}
  \cline{2-27}
   \multicolumn{1}{c|}{}       &\multicolumn{2}{c|}{aeroplane}& \multicolumn{2}{c|}{bicycle} & \multicolumn{2}{c|}{boat} & \multicolumn{2}{c|}{bottle} & \multicolumn{2}{c|}{bus} & \multicolumn{2}{c|}{car} & \multicolumn{2}{c|}{chair} & \multicolumn{2}{c|}{diningtable} & \multicolumn{2}{c|}{motorbike} & \multicolumn{2}{c|}{sofa} & \multicolumn{2}{c|}{train} & \multicolumn{2}{c||}{tvmonitor} & \multicolumn{2}{c|}{Avg.} \\ \hline

     4V  & 60.3\relax&49.0  & 52.8\relax&41.9  & 24.7\relax&12.8  & 21.6\relax&21.6  & 56.9\relax&52.3  & 46.5\relax&36.9  & 17.0\relax&11.8  & 26.1\relax&15.4  & 64.1\relax&53.1  & 26.3\relax&20.5  & 46.7\relax&38.5  & 54.8\relax&53.5  & 41.5\relax&33.9  \\ \hline 
     
     8V  & 61.3\relax&40.8  & 56.0\relax&34.7  & 18.9\relax&5.9  & 23.2\relax&23.2  & 58.0\relax&42.1  & 44.8\relax&30.7  & 14.8\relax&7.9  & 25.3\relax&12.8  & 56.7\relax&34.8  & 23.2\relax&13.3  & 42.8\relax&33.4  & 53.7\relax&40.9  & 39.9\relax&26.7  \\ \hline
     
     16V & 57.8\relax&18.4  & 53.1\relax&17.7  & 13.4\relax&2.1  & 21.5\relax&21.1  & 58.8\relax&47.4  & 43.4\relax&24.9  & 11.6\relax&3.9  & 23.5\relax&11.0  & 57.6\relax&23.5  & 19.9\relax&7.0  & 46.3\relax&31.9  & 50.7\relax&25.3  & 38.1\relax&19.5  \\ \hline
     
     24V & 54.7\relax&15.2  & 53.0\relax&11.0  & 16.5\relax&2.4  & 21.1\relax&20.5  & 58.1\relax&36.9  & 44.0\relax&18.5  & 11.6\relax&2.8  & 21.5\relax&7.4  & 54.4\relax&13.1  & 17.0\relax&5.3  & 44.1\relax&27.7  & 48.6\relax&22.7  & 37.0\relax&15.3  \\ \hline \hline
     24V I & 56.4\relax&18.2  & 49.9\relax&17.0  & 25.5\relax&4.0  & 23.0\relax&22.1  & 57.0\relax&37.8  & 41.0\relax&25.3  & 13.5\relax&4.5  & 29.8\relax&10.0  & 56.7\relax&20.1  & 33.7\relax&16.1  & 45.2\relax&31.0  & 48.1\relax&19.8  & 40.0\relax&18.8  \\ \hline
  \end{tabular}
  }
  \vspace{3mm}
  \caption{Single Class Continuous Regression - section \ref{sec:manifold}. The network aims at predicting points on a unit circle with the correct angle for positive instances and points far from the circle for negative instances.}
  \vspace{-2mm}
  \label{tab:manifold_all_classes}
  \centering
  {\tiny
  \begin{tabular}{|l@{~~}r| c | c | c | c | c | c | c | c | c | c | c | c || c |}
    \cline{3-15}
     \multicolumn{2}{c|}{}& aeroplane & bicycle & boat & bottle & bus & car & chair & diningtable & motorbike & sofa & train & tvmonitor & Avg. \\
   \hline
    \textbf{AP} &       & 53.9 & 49.9 & - & - & - & - & - & - & 52.7 & - & - & - & - \\ 
    \hline\hline
     \textbf{AVP} & 4V  & 45.0 & 41.7 & - & - & - & - & - & - & 42.0 & - & - & - & - \\ \hline
     \textbf{AVP} & 8V  & 37.5 & 32.1 & - & - & - & - & - & - & 30.4 & - & - & - & - \\ \hline
     \textbf{AVP} & 16V & 20.2 & 18.8 & - & - & - & - & - & - & 22.9 & - & - & - & -  \\ \hline
     \textbf{AVP} & 24V & 12.8 & 14.8 & - & - & - & - & - & - & 13.1 & - & - & - & -  \\ \hline 
  \end{tabular}
  }
  \vspace{3mm}
  \caption{Joint Continuous Pose Estimation for all classes - section \ref{sec:regression}. On the left side, only the pose-estimation is learned, the classifier is the baseline classifier. On the right side, both classification and pose-estimation are jointly trained.}
  \vspace{-2mm}
  \label{tab:regression_all}
  \centering
  {\tiny
  \begin{tabular}{|l@{~~}r|c|c|c||c||c|c|c|c|c|}
     \cline{3-9}
      \multicolumn{2}{c|}{} & \multicolumn{3}{c||}{$\lambda=0$ (fixed classifier)} & $\lambda=1$ &\multicolumn{3}{c|}{$\lambda=10$} \\[3pt] 
     \cline{3-9}
     \multicolumn{2}{c|}{} & \multicolumn{1}{c|}{$L^1$} & \multicolumn{1}{c|}{$L^2$} & \multicolumn{1}{c||}{Squared $L^2$} & \multicolumn{4}{c|}{$L^1$} \\[3pt]
     \cline{3-9}
     \multicolumn{2}{c|}{} & \multicolumn{3}{c||}{(b2)} & \multicolumn{1}{c||}{(b2)} & \multicolumn{1}{c|}{(a)} & \multicolumn{1}{c|}{(b1)} & \multicolumn{1}{c|}{(b2)}\\[3pt] 
     \hline
    \textbf{mAP}   &     & \multicolumn{3}{c||}{40.8}  & 37.8 & 43.0 & 38.9 & 44.1 \\ \hline \hline
     \textbf{mAVP} & 4V  & 31.6 &31.8  & 25.3 & 29.7 & 31.3 & 22.0 & 32.5 \\ \hline
     \textbf{mAVP} & 8V  & 24.3 &24.4  & 16.0 & 22.6 & 22.1 & 14.4 & 22.8 \\ \hline
    \textbf{mAVP}  & 16V & 17.9  &18.4  & 9.6 & 17.2 & 15.3 & 8.4 & 15.1 \\ \hline
    \textbf{mAVP}  & 24V & 14.3  &14.0  & 7.3 & 12.9 & 11.2 & 6.5 & 11.6 \\ \hline 
  \end{tabular}
  }
\end{table}


\paragraph{Continuous regression.}
\change{To study the method of \ref{sec:manifold} and determine the best set of parameters $(\lambda,\delta)$, we selected the bicycle class, for which all orientations are well represented in the database, and explored different values of the parameters (see table 
\ref{tab:manifold} in the appendix). 
We found that the model performs best for $\delta=1$ and $K=640$. However, we were not able to study yet higher values of $K$ because of stability issues. 
%
With this set of parameters, we conducted experiments on other classes. We report the results in table \ref{tab:manifold_all_classes}.\\
 The detection performances are better than the DPM baseline, but worse than the CNN baseline and the discrete method. This is not surprising since the error function is not optimized for detection.
The viewpoint predictions are clearly better than the CNN baseline, showing that pose information is effectively learned.}
 

\paragraph{Joint classification and continuous pose estimation.}
We first test the regression method \change{(b2)} without detection ($\lambda=0$) with different loss functions, using the same detector as in the CNN baseline to score the candidates. The results are presented on the left side of Table~\ref{tab:regression_all}. 
Using the squared $L^2$ norm as loss, we did not observe any improvement over the CNN baseline, showing that no real pose estimation is learned. The results obtained with $L^1$ and $L^2$ norms are \change{much} better, really learning pose estimation and \change{improving over both baselines. However, they are slightly lower than the results of the discrete method.}\\
We then investigate the interest of jointly optimizing detection and pose prediction. We test the three variants presented in section \ref{sec:regression} using the $L^ 1$ norm and $\lambda=10$, similar to ~\citet{penedones11}. The results are presented on the right side of Table~\ref{tab:regression_all}. The approach (b1) fails, probably because of its lack of flexibility. The most surprising result is the clear improvement of the detection performance for method (a) and (b2). This result corroborates the initial findings of \citet{penedones11} on a much more challenging dataset. A possible explanation is that separating the different orientations helps the network to find better representations of the object classes. Interestingly, despite this improvement in detection, the orientation predictions are not as good as those obtained by learning the pose independently from the detection. A possible explanation would be that the relative weight of the pose loss is too small. We thus repeated the experiment for method (b2) with $\lambda=1$. We observe only a limited improvement of the fine pose estimation performance, while the detection performance drops. \\

\paragraph{Summary.} Our experiments demonstrate that it is possible to perform joint detection and pose estimation with CNNs both in a continuous and discrete set-up. The discrete set-up provides the best results for orientation prediction, but it needs more data to avoid a decrease of the detection performance, because it treats each orientation as a separate class. On the contrary, it is possible in a continuous set-up to take advantage of the additional orientation annotations to improve detection performances.

\section{Conclusion}
In this paper, we presented a unified view of several CNN approaches for joint detection and orientation prediction. We compared these approaches and showed that there is a benefit in performing the two tasks jointly.  Finally, we used CNNs to outperform state-of-the-art results and provide a new baseline for joint detection and pose estimation.

\subsubsection*{Acknowledgments}
This work was carried out in IMAGINE, a joint research project between Ecole des Ponts ParisTech (ENPC) and the Scientific and Technical Centre for Building (CSTB). It was partly supported by ANR project Semapolis ANR-13-CORD-0003.

\setlength{\bibsep}{1pt plus 0.3ex}
\bibliography{iclr2015}

\begin{thebibliography}{28}
\providecommand{\natexlab}[1]{#1}
\providecommand{\url}[1]{\texttt{#1}}
\expandafter\ifx\csname urlstyle\endcsname\relax
  \providecommand{\doi}[1]{doi: #1}\else
  \providecommand{\doi}{doi: \begingroup \urlstyle{rm}\Url}\fi

\bibitem[Aubry et~al.(2014)Aubry, Maturana, Efros, Russell, and
  Sivic]{aubry2014seeing}
Aubry, M., Maturana, D., Efros, A.~A., Russell, B.~C., and Sivic, J.
\newblock Seeing {3D} chairs: exemplar part-based {2D-3D} alignment using a
  large dataset of {CAD} models.
\newblock In \emph{CVPR}, 2014.

\bibitem[Collobert et~al.(2011)Collobert, Kavukcuoglu, and
  Farabet]{torch7_NIPSworkshop2011}
Collobert, R., Kavukcuoglu, K., and Farabet, C.
\newblock Torch7: A {Matlab}-like environment for machine learning.
\newblock In \emph{BigLearn, NIPS Workshop}, 2011.
\newblock torch.ch.

\bibitem[Dalal \& Triggs(2005)Dalal and Triggs]{dalal2005histograms}
Dalal, N. and Triggs, B.
\newblock Histograms of oriented gradients for human detection.
\newblock In \emph{CVPR}, volume~1, pp.\  886--893, 2005.

\bibitem[Deng et~al.(2009)Deng, Dong, Socher, Li, Li, and
  Fei-Fei]{deng2009imagenet}
Deng, J., Dong, W., Socher, R., Li, L.-J., Li, K., and Fei-Fei, L.
\newblock {ImageNet}: A large-scale hierarchical image database.
\newblock In \emph{CVPR}, pp.\  248--255, 2009.

\bibitem[Everingham et~al.(2012)Everingham, Van~Gool, Williams, Winn, and
  Zisserman]{pascal-voc-2012}
Everingham, M., Van~Gool, L., Williams, C. K.~I., Winn, J., and Zisserman, A.
\newblock The {PASCAL} {V}isual {O}bject {C}lasses {C}hallenge 2012 {(VOC2012)}
  {R}esults.
\newblock
  http://www.pascal-network.org/challenges/VOC/voc2012/workshop/index.html,
  2012.

\bibitem[Fidler et~al.(2012)Fidler, Dickinson, and Urtasun]{fidler20123d}
Fidler, S., Dickinson, S., and Urtasun, R.
\newblock {3D} object detection and viewpoint estimation with a deformable {3D}
  cuboid model.
\newblock In \emph{Advances in Neural Information Processing Systems}, pp.\
  611--619, 2012.

\bibitem[Girshick et~al.(2013)Girshick, Donahue, Darrell, and
  Malik]{girshick2013rich}
Girshick, R., Donahue, J., Darrell, T., and Malik, J.
\newblock Rich feature hierarchies for accurate object detection and semantic
  segmentation.
\newblock \emph{preprint arXiv:1311.2524}, 2013.

\bibitem[He et~al.(2014)He, Zhang, Ren, and Sun]{kaiming14ECCV}
He, K., Zhang, X., Ren, S., and Sun, J.
\newblock Spatial pyramid pooling in deep convolutional networks for visual
  recognition.
\newblock In \emph{European Conf.\ on Comp.\ Vision (ECCV)}, pp.\  346--361,
  2014.

\bibitem[Huttenlocher \& Ullman(1990)Huttenlocher and
  Ullman]{huttenlocher1990recognizing}
Huttenlocher, D.~P. and Ullman, S.
\newblock Recognizing solid objects by alignment with an image.
\newblock \emph{IJCV}, 5\penalty0 (2):\penalty0 195--212, 1990.

\bibitem[Krizhevsky et~al.(2012)Krizhevsky, Sutskever, and
  Hinton]{krizhevsky2012imagenet}
Krizhevsky, A., Sutskever, I., and Hinton, G.~E.
\newblock {ImageNet} classification with deep convolutional neural networks.
\newblock In \emph{Advances in neural information processing systems}, pp.\
  1097--1105, 2012.

\bibitem[LeCun et~al.(1989)LeCun, Boser, Denker, Henderson, Howard, Hubbard,
  and Jackel]{lecun1989backpropagation}
LeCun, Y., Boser, B., Denker, J.~S., Henderson, D., Howard, R.~E., Hubbard, W.,
  and Jackel, L.~D.
\newblock Backpropagation applied to handwritten zip code recognition.
\newblock \emph{Neural comp.}, 1\penalty0 (4):\penalty0 541--551, 1989.

\bibitem[LeCun et~al.(1998)LeCun, Bottou, Bengio, and
  Haffner]{lecun1998gradient}
LeCun, Y., Bottou, L., Bengio, Y., and Haffner, P.
\newblock Gradient-based learning applied to document recognition.
\newblock \emph{Proceedings of the IEEE}, 86\penalty0 (11):\penalty0
  2278--2324, 1998.

\bibitem[Lim et~al.(2013)Lim, Pirsiavash, and Torralba]{lim2013parsing}
Lim, J.~J., Pirsiavash, H., and Torralba, A.
\newblock Parsing ikea objects: Fine pose estimation.
\newblock In \emph{International Conference on Computer Vision (ICCV)}, pp.\
  2992--2999. IEEE, 2013.

\bibitem[Lowe(2004)]{lowe2004distinctive}
Lowe, D.
\newblock Distinctive image features from scale-invariant keypoints.
\newblock \emph{IJCV}, 60\penalty0 (2):\penalty0 91--110, 2004.

\bibitem[Lowe(1987)]{lowe1987three}
Lowe, D.~G.
\newblock Three-dimensional object recognition from single two-dimensional
  images.
\newblock \emph{AIJ}, 31\penalty0 (3):\penalty0 355--395, 1987.

\bibitem[Mundy(2006)]{mundy2006object}
Mundy, J.~L.
\newblock Object recognition in the geometric era: A retrospective.
\newblock In \emph{Toward category-level object recognition}, pp.\  3--28.
  Springer, 2006.

\bibitem[Oquab et~al.(2014)Oquab, Bottou, Laptev, and Sivic]{Oquab13}
Oquab, M., Bottou, L., Laptev, I., and Sivic, J.
\newblock Learning and transferring mid-level image representations using
  convolutional neural networks.
\newblock In \emph{CVPR}, 2014.

\bibitem[Osadchy et~al.(2007)Osadchy, LeCun, and Miller]{osadchy07}
Osadchy, M., LeCun, Y., and Miller, M.~L.
\newblock Synergistic face detection and pose estimation with energy-based
  models.
\newblock \emph{The Journal of Machine Learning Research}, 8:\penalty0
  1197--1215, 2007.

\bibitem[Penedones et~al.(2011)Penedones, Collobert, Fleuret, and
  Grangier]{penedones11}
Penedones, H., Collobert, R., Fleuret, F., and Grangier, D.
\newblock Improving object classification using pose information.
\newblock Technical report, Idiap Research Institute, 2011.

\bibitem[Pepik et~al.(2012)Pepik, Stark, Gehler, and
  Schiele]{pepik2012teaching}
Pepik, B., Stark, M., Gehler, P., and Schiele, B.
\newblock Teaching {3D} geometry to deformable part models.
\newblock In \emph{CVPR}, pp.\  3362--3369, 2012.

\bibitem[Roberts(1963)]{roberts1963machine}
Roberts, L.~G.
\newblock \emph{Machine perception of three-dimensional solids}.
\newblock PhD thesis, Massachusetts Institute of Technology, 1963.

\bibitem[Sermanet et~al.(2013)Sermanet, Eigen, Zhang, Mathieu, Fergus, and
  LeCun]{sermanet2013overfeat}
Sermanet, P., Eigen, D., Zhang, X., Mathieu, M., Fergus, R., and LeCun, Y.
\newblock {OverFeat}: Integrated recognition, localization and detection using
  convolutional networks.
\newblock \emph{preprint arXiv:1312.6229}, 2013.

\bibitem[Simard et~al.(2003)Simard, Steinkraus, and Platt]{simard2003best}
Simard, P.~Y., Steinkraus, D., and Platt, J.~C.
\newblock Best practices for convolutional neural networks applied to visual
  document analysis.
\newblock In \emph{7th International Conference on Document Analysis and
  Recognition (ICDAR)}, volume~2, pp.\  958. IEEE Computer Society, 2003.

\bibitem[van~de Sande et~al.(2011)van~de Sande, Uijlings, Gevers, and
  Smeulders]{van2011segmentation}
van~de Sande, K. E.~A., Uijlings, J. R.~R., Gevers, T., and Smeulders, A. W.~M.
\newblock Segmentation as selective search for object recognition.
\newblock In \emph{ICCV}, pp.\  1879--1886. IEEE, 2011.

\bibitem[Williams et~al.(1997)Williams, Revow, and
  Hinton]{williams1997instantiating}
Williams, Christopher~KI, Revow, Michael, and Hinton, Geoffrey~E.
\newblock Instantiating deformable models with a neural net.
\newblock \emph{Computer Vision and Image Understanding}, 68\penalty0
  (1):\penalty0 120--126, 1997.

\bibitem[Xiang et~al.(2014)Xiang, Mottaghi, and Savarese]{xiang2014beyond}
Xiang, Y., Mottaghi, R., and Savarese, S.
\newblock Beyond {Pascal}: A benchmark for {3D} object detection in the wild.
\newblock In \emph{IEEE Winter Conference on Applications of Computer Vision
  (WACV)}, 2014.

\bibitem[Xiang \& Savarese(2012)Xiang and Savarese]{xiang2012estimating}
Xiang, Yu and Savarese, Silvio.
\newblock Estimating the aspect layout of object categories.
\newblock In \emph{Conference on Computer Vision and Pattern Recognition
  (CVPR)}, pp.\  3410--3417. IEEE, 2012.

\bibitem[Xiao et~al.(2012)Xiao, Russell, and Torralba]{xiao2012localizing}
Xiao, J., Russell, B., and Torralba, A.
\newblock Localizing {3D} cuboids in single-view images.
\newblock In \emph{Advances in Neural Information Processing Systems}, pp.\
  746--754, 2012.

\end{thebibliography}
\bibliographystyle{iclr2015}

\newpage
\appendix
\section{Supplementary Results}


\begin{table}[h!]
  \caption{Baseline - CNN + regression on fc7 features}
  \vspace{-3mm}
  \label{tab:linear}
  \centering
  \setlength{\tabcolsep}{5.7pt}
  {\tiny
  \begin{tabular}{|l@{~~}r| c | c | c | c | c | c | c | c | c | c | c | c || c |}
    \cline{3-15}
     \multicolumn{2}{c|}{}   & aeroplane & bicycle & boat & bottle & bus & car & chair & diningtable & motorbike & sofa & train & tvmonitor & Avg. \\
 \hline
   \textbf{AP}  &  & 58.9 & 53.8 & 24.8 & 23.3 & 56.9 & 44.6 & 16.6 & 28.3 & 57.4 & 27.6 & 42.3 & 54.8 & 40.8 \\
    \hline\hline
     \textbf{AVP} & 4V  & 19.9 & 19.7 & 9.3 & 23.2 & 47.2 & 20.7 & 6.0 & 16.4 & 15.8 & 22.4 & 35.5 & 52.4 & 23.9 \\ \hline
     \textbf{AVP} & 8V  & 9.7 & 10.6 & 2.5 & 23.2 & 28.7 & 10.5 & 2.9 & 10.2 & 4.6 & 11.8 & 27.3 & 38.1 & 14.9 \\ \hline
     \textbf{AVP} & 16V & 4.5 & 4.2 & 1.1 & 22.9 & 9.7 & 5.0 & 1.6 & 9.5 & 2.1 & 7.8 & 14.1 & 22.6 & 8.7  \\ \hline
     \textbf{AVP} & 24V & 3.8 & 4.0 & 0.3 & 22.4 & 6.5 & 4.2 & 1.4 & 8.2 & 0.9 & 6.8 & 8.9 & 16.0 & 6.9  \\ \hline 
  \end{tabular}
  }
\end{table}

\begin{table}[h!]
  \caption{Baseline - CNN + Most probable viewpoint}
  \vspace{-3mm}
  \label{tab:maxpresence}
  \centering
  \setlength{\tabcolsep}{5.7pt}
  {\tiny
  \begin{tabular}{|l@{~~}r| c | c | c | c | c | c | c | c | c | c | c | c || c |}
    \cline{3-15}
     \multicolumn{2}{c|}{}   & aeroplane & bicycle & boat & bottle & bus & car & chair & diningtable & motorbike & sofa & train & tvmonitor & Avg. \\
 \hline
   \textbf{AP}  &  & 58.9 & 53.8 & 24.8 & 23.3 & 56.9 & 44.6 & 16.6 & 28.3 & 57.4 & 27.6 & 42.3 & 54.8 & 40.8 \\
    \hline\hline
     \textbf{AVP} & 4V  & 17.8 & 23.5 & 6.6 & 23.2 & 47.2 & 10.5 & 6.1 & 16.2 & 21.6 & 22.5 & 35.6 & 52.8 & 23.6 \\ \hline
     \textbf{AVP} & 8V  & 10.3 & 8.0 & 5.2 & 23.2 & 28.8 & 7.0 & 2.9 & 10.2 & 13.8 & 12.0 & 27.6 & 38.4 & 15.6 \\ \hline
     \textbf{AVP} & 16V & 6.4 & 5.5 & 2.0 & 22.9 & 18.9 & 3.6 & 1.2 & 9.5 & 7.1 & 8.0 & 10.7 & 17.0 & 9.4  \\ \hline
     \textbf{AVP} & 24V & 4.1 & 3.8 & 1.0 & 22.5 & 11.4 & 2.5 & 0.9 & 8.1 & 5.7 & 6.9 & 10.8 & 18.0 & 8.0  \\ \hline 
  \end{tabular}
  }
\end{table}

\begin{table}[h!]
  \caption{Single-Class Continuous regression - section \ref{sec:manifold}. Grid search for $K$ and $\delta$}
  \vspace{-3mm}
  \label{tab:manifold}
  \centering
  {\tiny
  \begin{tabular}{|l@{~~}r|c|c|c|c|c|c|c|c|c|c|c|}
     \cline{3-11}
     \multicolumn{2}{c|}{} &\multicolumn{9}{c|}{ bicycle}\\
     \cline{3-11}
     \multicolumn{2}{c|}{} & \multicolumn{1}{c|}{$\delta=0.25$} & \multicolumn{1}{c|}{$\delta=0.5$} & \multicolumn{1}{c|}{$\delta=1$} & \multicolumn{3}{c|}{$\delta=2$} & \multicolumn{3}{c|}{ $\delta=4$} \\
     \cline{3-11}
     \multicolumn{2}{c|}{} & K=640 & K=640 & K=640 & K=160 & K=320 & K=640 & K=160 & K=320  & K=640\\ 
     \hline     
    \textbf{AP}   &     & 23.8 & 45.8 & \bf 49.9 &34.9  & 41.8 & 47.4 & 32.5 & 40.9 & 43.4 \\ \hline \hline
     \textbf{AVP} & 4V  & 16.0 & 38.8 & \bf 41.7 &29.3  & 35.8 & 39.7 & 28.0 & 35.8 & 36.3\\ \hline
     \textbf{AVP} & 8V  & 11.6 & 28.3 &\bf  32.1 &22.7  & 25.7 & 27.7 & 21.6 &26.5  & 27.0 \\ \hline
    \textbf{AVP}  & 16V & 4.9 & 14.5 &\bf  18.8  &12.9  & 14.8 & 17.2 & 10.4 & 14.5 & 14.9 \\ \hline
    \textbf{AVP}  & 24V & 3.7 & 13.0 & \bf 14.8  &9.5  & 13.1 & 14.6 & 8.1  & 13.0  & 12.7 \\ \hline 
  \end{tabular}
  }
\end{table}

\begin{table}[h!]
  \caption{Fixed Classifier - Regression with $\lambda=0$ - section \ref{sec:regression} - variant  (b2), $L^1$ norm}
  \vspace{-3mm}
  \label{tab:regression_l1}
  \centering
  {\tiny
  \begin{tabular}{|l@{~~}r| c | c | c | c | c | c | c | c | c | c | c | c || c |}
    \cline{3-15}
     \multicolumn{2}{c|}{}   & aeroplane & bicycle & boat & bottle & bus & car & chair & diningtable & motorbike & sofa & train & tvmonitor & Avg. \\
   \hline
    \textbf{AP} & &  58.9 & 53.8 & 24.8 & 23.3 & 56.9 & 44.6 & 16.6 & 28.3 & 57.4 & 27.6 & 42.3 & 54.8 & 40.8 \\ 
    \hline\hline
     \textbf{AVP} & 4V  & 43.7 & 41.1 & 8.8 & 23.2 & 48.9 & 29.7 & 9.7 & 16.4 & 46.9 & 21.9 & 35.9 & 53.3 & 31.6 \\ \hline
     \textbf{AVP} & 8V  & 33.3 & 29.8 & 6.1 & 23.2 & 37.2 & 23.2 & 7.6 & 11.6 & 33.8 & 16.0 & 29.3 & 40.3 & 24.3 \\ \hline
     \textbf{AVP} & 16V & 19.3 & 17.1 & 4.6 & 22.9 & 38.4 & 16.9 & 5.1 & 6.3 & 24.3 & 10.0 & 23.1 & 26.5 & 17.9  \\ \hline
     \textbf{AVP} & 24V & 13.9 & 14.2 & 2.7 & 22.5 & 29.3 & 14.8 & 3.5 & 6.6 & 17.7 & 8.2 & 17.6 & 20.3 & 14.3  \\ \hline 
  \end{tabular}
  }
\end{table}

\begin{table}[h!]
  \caption{Fixed Classifier - Regression with $\lambda=0$ - section \ref{sec:regression} - variant  (b2), $L^2$ norm}
  \vspace{-3mm}
  \label{tab:regression_l2}
  \centering
  {\tiny
  \begin{tabular}{|l@{~~}r| c | c | c | c | c | c | c | c | c | c | c | c || c |}
    \cline{3-15}
     \multicolumn{2}{c|}{}   & aeroplane & bicycle & boat & bottle & bus & car & chair & diningtable & motorbike & sofa & train & tvmonitor & Avg. \\
   \hline
    \textbf{AP} & &  58.9 & 53.8 & 24.8 & 23.3 & 56.9 & 44.6 & 16.6 & 28.3 & 57.4 & 27.6 & 42.3 & 54.8 & 40.8 \\ 
    \hline\hline
     \textbf{AVP} & 4V  & 43.8 & 40.6 & 10.3 & 23.2 & 48.9 & 31.1 & 9.4 & 17.1 & 46.1 & 21.4 & 35.8 & 53.4 & 31.8 \\ \hline
     \textbf{AVP} & 8V  & 33.2 & 30.6 & 8.0 & 23.2 & 39.0 & 24.1 & 7.5 & 9.8 & 31.7 & 15.9 & 29.4 & 39.8 & 24.4 \\ \hline
     \textbf{AVP} & 16V & 22.5 & 15.7 & 3.7 & 22.9 & 41.0 & 18.2 & 5.0 & 6.7 & 20.1 & 12.8 & 23.7 & 28.6 & 18.4  \\ \hline
     \textbf{AVP} & 24V & 12.1 & 11.0 & 3.3 & 22.5 & 32.1 & 15.4 & 4.0 & 3.3 & 16.2 & 7.5 & 19.9 & 20.0 & 14.0  \\ \hline 
  \end{tabular}
  }
\end{table}

\begin{table}[h!]
  \caption{Fixed Classifier - Regression with $\lambda=0$ - section~\ref{sec:regression} - variant (b2), Squared $L^2$ norm}
  \vspace{-3mm}
  \label{tab:regression_l22}
  \centering
  {\tiny
  \begin{tabular}{|l@{~~}r| c | c | c | c | c | c | c | c | c | c | c | c || c |}
    \cline{3-15}
     \multicolumn{2}{c|}{}   & aeroplane & bicycle & boat & bottle & bus & car & chair & diningtable & motorbike & sofa & train & tvmonitor & Avg. \\
   \hline
    \textbf{AP} & &  58.9 & 53.8 & 24.8 & 23.3 & 56.9 & 44.6 & 16.6 & 28.3 & 57.4 & 27.6 & 42.3 & 54.8 & 40.8 \\ 
    \hline\hline
     \textbf{AVP} & 4V  & 36.4 & 12.4 & 9.0 & 23.2 & 48.0 & 23.3 & 7.6 & 16.2 & 16.1 & 22.5 & 35.6 & 52.8 & 25.3 \\ \hline
     \textbf{AVP} & 8V  & 21.0 & 6.9 & 2.8 & 23.2 & 28.5 & 12.2 & 4.1 & 10.2 & 5.2 & 12.0 & 27.6 & 38.4 & 16.0 \\ \hline
     \textbf{AVP} & 16V & 15.1 & 3.3 & 1.1 & 22.9 & 9.5 & 5.6 & 2.4 & 8.7 & 2.4 & 8.0 & 14.3 & 22.4 & 9.6  \\ \hline
     \textbf{AVP} & 24V & 7.6 & 3.6 & 1.5 & 22.5 & 6.5 & 4.5 & 1.8 & 5.6 & 1.4 & 6.8 & 9.6 & 16.0 & 7.3  \\ \hline 
  \end{tabular}
  }
\end{table}

\begin{table}[h!]
  \caption{Joint Classification and continuous pose estimation for all classes with $\lambda=1$ - section~\ref{sec:regression} - variant  (b2), $L^1$ norm}
  \vspace{-3mm}
  \label{tab:shared_l1_trial2}
  \centering
  {\tiny
  \begin{tabular}{|l@{~~}r| c | c | c | c | c | c | c | c | c | c | c | c || c |}
    \cline{3-15}
     \multicolumn{2}{c|}{}   & aeroplane & bicycle & boat & bottle & bus & car & chair & diningtable & motorbike & sofa & train & tvmonitor & Avg. \\
   \hline
    \textbf{AP} & &  52.0 & 46.1 & 22.4 & 19.0 & 51.3 & 38.8 & 19.4 & 26.1 & 61.3 & 30.6 & 39.0 & 47.3 & 37.8 \\ 
    \hline\hline
     \textbf{AVP} & 4V  & 40.0 & 34.2 & 9.9 & 18.9 & 45.4 & 26.6 & 12.6 & 14.6 & 51.1 & 24.5 & 33.5 & 44.8 & 29.7 \\ \hline
     \textbf{AVP} & 8V  & 30.3 & 27.6 & 6.3 & 18.9 & 30.6 & 20.9 & 9.7 & 10.9 & 37.4 & 17.7 & 27.3 & 33.8 & 22.6 \\ \hline
     \textbf{AVP} & 16V & 15.3 & 15.9 & 5.6 & 18.6 & 34.4 & 15.3 & 6.3 & 8.1 & 26.6 & 15.6 & 21.1 & 23.1 & 17.2  \\ \hline
     \textbf{AVP} & 24V & 10.8 & 13.1 & 3.1 & 18.1 & 25.5 & 11.2 & 4.1 & 5.4 & 20.8 & 9.5 & 16.6 & 16.1 & 12.9  \\ \hline 
  \end{tabular}
  }
\end{table}

\begin{table}[h!]
  \caption{Joint Classification and continuous pose estimation for all classes with $\lambda=10$ - section~\ref{sec:regression} - variant  (a), $L^1$ norm}
  \vspace{-3mm}
  \label{tab:shared_l1_lambda10_a}
  \centering
  {\tiny
  \begin{tabular}{|l@{~~}r| c | c | c | c | c | c | c | c | c | c | c | c || c |}
    \cline{3-15}
     \multicolumn{2}{c|}{}   & aeroplane & bicycle & boat & bottle & bus & car & chair & diningtable & motorbike & sofa & train & tvmonitor & Avg. \\
   \hline
    \textbf{AP} & &  61.6 & 58.7 & 26.3 & 21.7 & 58.3 & 45.4 & 23.3 & 27.7 & 64.3 & 30.8 & 45.7 & 52.3 & 43.0 \\ 
    \hline\hline
     \textbf{AVP} & 4V  & 45.3 & 36.3 & 9.8 & 21.7 & 48.2 & 24.2 & 13.3 & 14.8 & 48.5 & 24.5 & 38.8 & 50.4 & 31.3 \\ \hline
     \textbf{AVP} & 8V  & 27.2 & 26.3 & 4.0 & 21.7 & 31.9 & 16.7 & 9.0 & 8.9 & 35.3 & 15.3 & 33.4 & 36.1 & 22.1 \\ \hline
     \textbf{AVP} & 16V & 19.6 & 14.4 & 3.3 & 21.4 & 23.7 & 11.2 & 6.0 & 7.1 & 21.3 & 10.3 & 22.5 & 22.8 & 15.3  \\ \hline
     \textbf{AVP} & 24V & 11.3 & 10.1 & 1.1 & 21.0 & 18.1 & 9.2 & 3.5 & 5.1 & 15.5 & 8.6 & 18.2 & 12.1 & 11.2  \\ \hline 
  \end{tabular}
  }
\end{table}

\begin{table}[h!]
  \caption{Joint Classification and continuous pose estimation for all classes with $\lambda=10$ - section~\ref{sec:regression} - variant  (b1), $L^1$ norm}
  \vspace{-3mm}
  \label{tab:shared_l1_lambda10_b1}
  \centering
  {\tiny
  \begin{tabular}{|l@{~~}r| c | c | c | c | c | c | c | c | c | c | c | c || c |}
    \cline{3-15}
     \multicolumn{2}{c|}{}   & aeroplane & bicycle & boat & bottle & bus & car & chair & diningtable & motorbike & sofa & train & tvmonitor & Avg. \\
   \hline
    \textbf{AP} & &  53.2 & 55.1 & 20.9 & 22.0 & 53.7 & 42.3 & 20.2 & 27.9 & 56.2 & 30.1 & 36.1 & 49.2 & 38.9 \\ 
    \hline\hline
     \textbf{AVP} & 4V  & 15.8 & 15.0 & 7.9 & 21.9 & 46.4 & 18.7 & 6.7 & 14.0 & 17.7 & 23.6 & 29.4 & 47.3 & 22.0 \\ \hline
     \textbf{AVP} & 8V  & 11.1 & 8.9 & 3.2 & 21.9 & 30.2 & 8.5 & 2.9 & 9.0 & 6.0 & 13.9 & 22.4 & 34.2 & 14.4 \\ \hline
     \textbf{AVP} & 16V & 5.1 & 3.6 & 1.1 & 21.5 & 9.7 & 4.2 & 1.5 & 8.4 & 4.6 & 9.8 & 10.8 & 20.4 & 8.4  \\ \hline
     \textbf{AVP} & 24V & 4.7 & 1.8 & 0.9 & 21.2 & 6.8 & 3.5 & 1.2 & 6.9 & 3.3 & 8.3 & 6.2 & 13.4 & 6.5  \\ \hline 
  \end{tabular}
  }
\end{table}

\begin{table}[h!]
  \caption{Joint Classification and continuous pose estimation for all classes with $\lambda=10$ section \ref{sec:regression} - variant  (b2), $L^1$ norm}
  \vspace{-3mm}
  \label{tab:shared_l1_lambda10_b2}
  \centering
  {\tiny
  \begin{tabular}{|l@{~~}r| c | c | c | c | c | c | c | c | c | c | c | c || c |}
    \cline{3-15}
     \multicolumn{2}{c|}{}   & aeroplane & bicycle & boat & bottle & bus & car & chair & diningtable & motorbike & sofa & train & tvmonitor & Avg. \\
   \hline
    \textbf{AP} & &  63.1 & 59.4 & 26.4 & 23.7 & 57.7 & 46.5 & 22.4 & 29.9 & 64.6 & 35.6 & 46.3 & 53.9 & 44.1 \\ 
    \hline\hline
     \textbf{AVP} & 4V  & 43.4 & 37.7 & 8.9 & 23.6 & 50.5 & 26.0 & 13.0 & 15.8 & 51.3 & 29.9 & 38.9 & 51.3 & 32.5 \\ \hline
     \textbf{AVP} & 8V  & 31.2 & 24.4 & 5.3 & 23.6 & 33.1 & 16.9 & 9.3 & 8.7 & 31.9 & 20.5 & 30.9 & 37.5 & 22.8 \\ \hline
     \textbf{AVP} & 16V & 18.5 & 13.8 & 3.7 & 23.4 & 22.8 & 11.5 & 5.7 & 7.2 & 19.2 & 12.8 & 19.8 & 23.0 & 15.1  \\ \hline
     \textbf{AVP} & 24V & 10.7 & 11.3 & 2.3 & 22.9 & 13.2 & 7.0 & 3.9 & 4.3 & 15.6 & 10.7 & 21.0 & 15.9 & 11.6  \\ \hline 
  \end{tabular}
  }
\end{table}

\end{document}